\documentclass[runningheads]{llncs}
\usepackage[T1]{fontenc}
\usepackage{graphicx}
\usepackage{algorithm}
\usepackage{algorithmic}
\usepackage{subfigure}
\usepackage{multirow}
\usepackage{makecell}
\usepackage{amsmath,amssymb} 
\usepackage{booktabs}
\usepackage{multirow}
\usepackage{url}
\usepackage{subcaption}
\usepackage{array}
\usepackage{graphicx}
\usepackage{enumitem} % 导入宏包  itemize 列表
\usepackage{colortbl}  %彩色表格需要加载的宏包
\usepackage{xcolor}
\usepackage{soul} % 导入 soul 包
\usepackage{color, xcolor} 
\usepackage{tabularx}
\newcolumntype{L}[1]{>{\raggedright\let\newline\\\arraybackslash\hspace{0pt}}m{#1}}
\newcolumntype{C}[1]{>{\centering\let\newline\\\arraybackslash\hspace{0pt}}m{#1}}
\newcolumntype{R}[1]{>{\raggedleft\let\newline\\\arraybackslash\hspace{0pt}}m{#1}}
\usepackage{enumerate}
\usepackage{float}
\usepackage{soul}

\usepackage{color, xcolor} 
\usepackage[colorlinks,
            linkcolor=blue,
            anchorcolor=blue,
            citecolor=blue]{hyperref}

\usepackage{footmisc}

\begin{document}
\newsavebox\CBox
\def\textBF#1{\sbox\CBox{#1}\resizebox{\wd\CBox}{\ht\CBox}{\textbf{#1}}}

\title{Mitigating Hallucination in Visual-Language Models via Re-Balancing Contrastive Decoding}

\author{Xiaoyu Liang\inst{1}\thanks{These authors contributed equally to this work.} \and Jiayuan Yu \inst{1}$^{*}$ 
\and 
Lianrui  Mu \inst{1} \and Jiedong Zhuang\inst{1} \and
Jiaqi Hu\inst{1} \and Yuchen Yang\inst{1} \and  Jiangnan Ye\inst{1} \and Lu Lu\inst{2} \and Jian Chen\inst{2} \and Haoji Hu\inst{1}\thanks{Corresponding author.}}

\authorrunning{Xiaoyu. et al.}

\institute{College of Information Science and Electronic Engineering, Zhejiang University, China \and Alibaba Group
\\
\email{\{xiaoyu\_l, jiayuan\_yu, mulianrui, zhuangjiedong, jiaqi\_hu,\\
yuchen\_yang, jiangnan\_ye, haoji\_hu\}@zju.edu.cn,\\
\{ll200214,j.chen
\}@alibaba-inc.com}}
\maketitle

\begin{abstract} %RBD 
Although Visual-Language Models (VLMs) have shown impressive capabilities in tasks like visual question answering and image captioning, they still struggle with hallucinations. Analysis of attention distribution in these models shows that VLMs tend to processing textual tokens rather than visual tokens. This imbalance of attention distribution causes VLMs to favor textual knowledge in the case of multimodal knowledge conflicts, resulting in differences from the image information. In this paper, we propose Re-Balancing Contrastive Decoding (RBD) method, which employs textual and visual branches to recalibrate attention distribution in VLMs. Specifically, the textual branch injects image noise to stimulate the model's dependency on text, thereby reducing textual bias. Concurrently, the visual branch focuses on the selection of significant tokens, refining the attention mechanism to highlight the primary subject. This dual-branch strategy enables the RBD method to diminish textual bias while enhancing visual information. Experimental results demonstrate that our method, RBD, outperforms the existing methods by the CHAIR and POPE metrics, mitigate hallucinations without reducing the model's general capabilities. 

\end{abstract}

\section{Introduction}
\label{sec_Intro}

Visual-Language Models (VLMs)~\cite{bai2023qwen,liu2023llavaplus,liu2023llava} leverage the advanced capabilities of Large Language Models (LLMs) to yield text responses that in harmony with the content of the input images. 
By combining image understanding with natural language processing, VLMs can tackle various tasks such as visual question answering, image captioning, and object localization, marking a significant advancement in the field of multimodal intelligence.

\begin{figure}[!t]
\includegraphics[width=\textwidth]{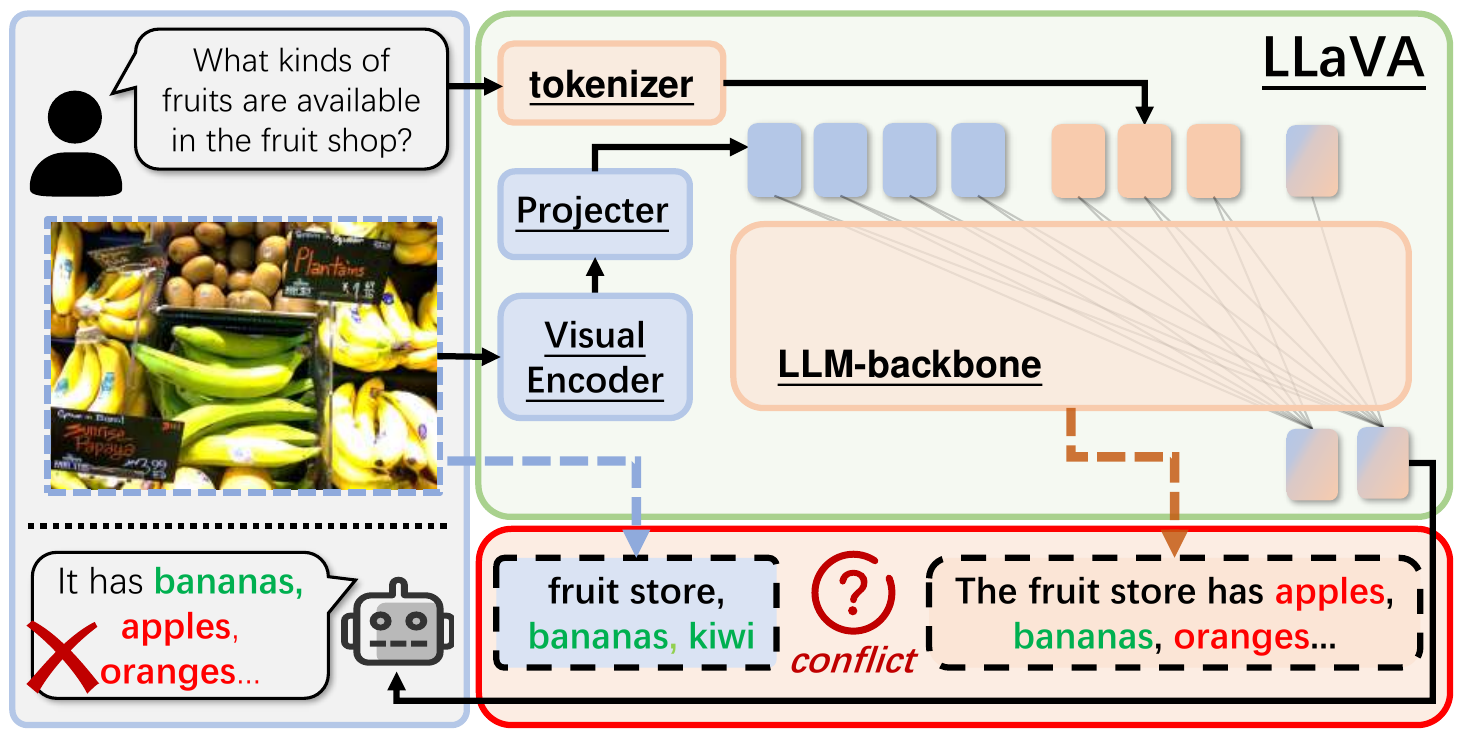}
\caption{
\textbf{Imbalance in Multimodal Knowledge Processing.} 
\underline{LLaVA} tends to processing \colorbox[RGB]{248,203,173}{textual} rather than \colorbox[RGB]{180,199,231}{visual} information. 
LLaVAv1.5 assume the presence of apples in a fruit shop, even if there is no apple in the image. This assumption is influenced by the inherent textual knowledge stored in the LLM-backbone, thereby creating hallucinations.
Words marked in \textcolor{red}{red} and \textcolor{green}{green} show incorrect and correct information, respectively. 
} \label{figfirstshow}
\end{figure}

Nevertheless, in mainstream VLM architectures, a lightweight visual encoder is paired with a heavyweight and powerful LLM backbone through a cross-modality projector. 
An imbalance in the visual and textual components of VLM, or inadequate alignment between them, can lead to an excessive reliance on textual data~\cite{wang2024cogvlm}. 
Furthermore, analysis of the attention distribution (Sec.~\ref{sec_attention}) further validates that imbalance. 
As illustrated in Fig.~\ref{figfirstshow}, when there is a conflict within modalities knowledge, the golden information of the image will be ignored, 
which is called \textit{Multimodal Knowledge Conflicting Hallucinations}.

Prior researches have primarily attributed hallucinations to defects in the visual encoder or insufficient modality alignment, and attempted to mitigate hallucinations by fine-tuning~\cite{yu2024rlhfv,zhao2024hallucinations,gunjal2024detecting,pi2024strengthening}, integrating external tools~\cite{deng2024seeing,zhou2024analyzing} and contrastive decoding~\cite{chuang2024dola,zhang2024alleviating,leng2023mitigating}. 
However, fine-tuning requires building high-quality datasets and consumes significant computing resources, introducing external tools alters the initial model's output, potentially diverging from user instructions. 
Recently, contrastive decoding methods have gained popularity for their elegant simplicity and effectiveness. Methods such as CD~\cite{zhang2024alleviating}, DoLa~\cite{chuang2024dola} and VCD~\cite{leng2023mitigating} have shown efficacy by modifying logits. However, these methods not fully resolved the issue of the imbalanced biases in VLMs.

Given that the input images are considered as ``gold standard'', 
we prefer that the VLM's responses rely on visual information rather than speculation or unsupported elaboration. 
We propose our method, \textbf{R}e-\textbf{B}alancin Contrastive \textbf{D}ecoding (\textbf{RBD}). 
Fig.~\ref{figoverall} provides a overview of our method. 
Our RBD method, inspired by the VCD~\cite{leng2023mitigating}, employs textual and visual branches to recalibrate attention distribution in VLMs. 
We employ the visual branch to amplify significant visual tokens, while utilizing the textual branch to identify tokens originate from inherent textual knowledge, subsequently penalizing these tokens to mitigate the influence of textual bias. 
Experimental results~\ref{sec:exp} validate the effectiveness of our proposed RBD method, which surpasses current methods in CHAIR and POPE metrics, minimizing hallucinations while maintaining VLM's overall efficacy. 
Our contributions are summarised as follows:
\begin{itemize}[leftmargin=*,nosep]
    \item[$\star$] Analysis of attention distribution reveals a model bias towards text over images, suggesting a new avenue for exploring model hallucinations.
    
    \item[$\star$] Our method RBD adopt two auxiliary branches to recalibrate the VLM's dependency on visual and textual information. 

    \item[$\star$] Our method, as a plug-and-play method, achieves superior performance compared with other state-of-the-art methods in CHAIR and POPE metrics. 
\end{itemize}

\section{Related Work}

\subsection{Vision-Language Hallucination}

VLMs derived from LLMs exhibit vision-language hallucination~\cite{li2023evaluating}. Recent studies have begun to investigate the issue of object hallucination~\cite{li2023evaluating} extensively, initially focusing on object existence and gradually expanding to finer-grained errors, including object attributes, spatial relationships, physical states, activities, and numerical inaccuracies~\cite{gunjal2024detecting,wang2024amber,guan2024hallusionbench,liu2024phd}.
Sources of vision-language hallucinations: 
(1) Lack of fine-grained representation or spatial information, due to limited image resolution~\cite{ai2024yi,li2024monkey,li2023otterhd}; 
(2) Lack of cross-modality representation alignment~\cite{wang2024cogvlm,jian2024expedited,li2022supervision,kim2021vilt,liu2023llavaplus}; 
(3) The tendency of LLMs to hallucinate, because their overconfident about parameter-driven internal knowledge. And errors will accumulate over auto-regressive decoding processes, ultimately resulting in hallucinations~\cite{zhang2023sirens,mckenna2023sources}.

We aim to examine the essence of hallucinations, positioning them as Multimodal knowledge conflicting hallucinations, which highlights the VLM's tendency to favor its internal textual knowledge over external visual information, particularly when they are faced with conflicting information.

\subsection{Mitigating Vision-Language Hallucination}
\label{sec-3ways2maitigating}

Previous research has explored some approaches to mitigate hallucinations in LVLMs, which can be categorized into three main groups:

\textbf{Utilization of External Tools. }
Utilizing Optical Character Recognition (OCR) or segmentation models is a significant strategy. Woodpecker~\cite{yin2023woodpecker}, a groundbreaking research initiative addressing hallucination issues in VLMs, utilizes GroundingDINO~\cite{liu2023grounding} for target detection. By leveraging post-hoc correctors based on other models, corrections can result in outputs with reduced hallucinations~\cite{deng2024seeing,zhou2024analyzing}. An example is CGD~\cite{deng2024seeing}, which employs CLIP~\cite{radford2021learning} to evaluate the accuracy of generated sentences.

\textbf{Training Interventions. }
Hallucinations can be mitigated  by training on a more detailed or diverse set of instruction data, such as COG-VLM~\cite{wang2024cogvlm} and Qwen-VL~\cite{bai2023qwen}.
Human feedback reinforcement learning~\cite{yu2024rlhfv,zhao2024hallucinations,gunjal2024detecting,pi2024strengthening}, strengthened with fact augmentation, has also shown to be effective. 
While these methods are effective, they require complex training strategies and carefully constructed data datasets to tune model parameters, the labor and computational expenses involved hinder their large-scale implementation.

\textbf{Decoding Strategies. }
Without necessitating additional training, decoding strategies have been developed to alleviate hallucinations in multimodal large models. 
In the field of LLMs, notable strategies like DoLa~\cite{chuang2024dola} and ITI~\cite{li2023inferencetime} may provide insights for addressing hallucinations. 
OPERA~\cite{huang2024opera} provides an in-depth analysis of the mechanism of hallucinations, deploying attention penalization and fallback strategies during decoding. 
VCD~\cite{leng2023mitigating}, modifies the model's output distribution logits to reduce hallucinations. Likewise, decoding strategies like ICD~\cite{zhang2024alleviating}, and IBD~\cite{zhu2024ibd} have been introduced to address this issue.

\section{Method}

\subsection{Preliminary}

\subsubsection{Architecture.}

Fig.~\ref{figfirstshow} demonstrates the key elements of the VLMs architecture: a Visual Encoder, a LLM Backbone and a Cross-modality Projector. 
Serving as a pivotal bridge between modalities, the Cross-modality Projector maps visual vectors into the textual vector space, and is typically structured around one of three core mechanisms: cross-attention, Q-Former, or MLP~\cite{li2024llava}. 

\subsubsection{Decoding.}

We conceptualize VLM as a unified entity with parameters denoted as $\theta$. 
For given image-instruction pair $( v,x )$, 
VLM combines transformed visual and textual tokens into an input sequence provided to the LLM Backbone for iterative next token prediction until encountering a termination token. 
The mathematical formulation of the process is as follows:
\begin{equation}
y_t \sim P_{\theta}(y_t|v,x,y_{<t}) = Softmax( logit_{\theta}(y_t |v,x,y_{<t}))
\end{equation}

\noindent Where $logit_{\theta}(\cdot)$ donates 
the function that computes the unnormalized prediction scores by VLM ${\theta}$ for a given input, 
and at each time step \(t\), the decoding process computing the logits using the prior tokens \(y_{<t}\), the input sequence \( v \) and \( x\), applying the softmax function to obtain probabilities.

\subsubsection{Contrastive Decoding.}

Through contrastive decoding, we can modify the model’s logits to regulate the preferences of the VLM model. 
More specifically, we construct two contrastive branches: the standard and the conditional. Both branches undergo forward inference independently, producing different logits. Decoding is then based on their discrepancies, as shown in the following equation:
\begin{equation}
P_{CD}(y|v,x,s) = Softmax[(1-\alpha)logit_{\theta}(y |v,x) + \alpha logit_{\theta}(y|v,x,s)]
\end{equation}

\noindent Where $\text{logit}_{\theta}(y|v,x,s)$ represents the logits from the conditional branch, with $s$ denoting operations that differentiate from the standard branch. These operations include, but are not limited to, altering model parameters, modifying the input image, or changing the input text~\cite{zhang2024alleviating,leng2023mitigating,chuang2024dola}.
And we also introduce an additional hyperparameter $\alpha \in [0,1]$ to control the intensity of the contrastive mechanism. Consequently, setting $\alpha$ to 0 implies that decoding is exclusively reliant on the original standard branch.

Next, we utilize the distribution $P_{CD}(y|v,x,s)$ for the next token $y_t$ prediction, by sampling from these probabilities or selecting the most probable token.

\subsection{Analysis of Attention Distribution}
\label{sec_attention} 
\iffalse

\begin{figure}[!t]
\includegraphics[width=\textwidth]{stacked_area_chart_240424.png}
\caption{

Attention distribution stack plots for four types of tokens during the decoding process of LLaVAv1.5-7B, with each column representing the attention distribution within a layer. 
} \label{figattention}

\end{figure}
Fig.~\ref{figattention} shows the attention distribution within the LLaVAv1.5-7B model, revealing an imbalance where the attention is more on textual tokens than on visual tokens. 
\fi

The attention distribution within the LLaVAv1.5-7B model, revealing an imbalance where the attention is more on textual tokens than on visual tokens. 
During the decoding phase, attention maps are gathered from each layer, denoted as $A_k \in \mathbb{R}^{n \times n} $. For this process, we randomly selected a subset of 500 image-instruction pairs $(v,x)$  from LLaVA-Instruct-150K dataset. 
We categorized input tokens into system (\(sys\)), image (\(img\)), instruction (\(ins\)), and response (\(res\)) types, and computed the cumulative attention weights for each token type, maintaining the sum of attention weights satisfies the following equation:

\begin{equation}
A_{k,i}^{sys}+A_{k,i}^{img}+A_{k,i}^{ins}+A_{k,i}^{res} = 1
\end{equation}
\noindent Where $A_{k,i}^{sys}$ donates the attention that token $i$ pays to system token: 
\begin{equation}
A_{k,i}^{sys} = \sum_{j\in \{sys\}} A_{k}^{i,j}
\end{equation}
The elements  at row $i$ and column $j$ of the attention matrix $A_k$, denoted as $A_{k}^{i,j}$, represent the attention that token $i$ pays to token $j$. And we set $i$ to the tenth token generated by LLaVAv1.5.

In the the deep layers, attention scores are predominantly concentrated on `sys' and `res' tokens, while `img' tokens are largely disregarded. This phenomenon indirectly supports cogvlm's perspective on the insufficiency of deep multimodal integration in VLMs~\cite{wang2024cogvlm}. 
Specifically, visual tokens receive merely 25$\%$ of the total attention, and this imbalance intensifies in deeper layers.  
This imbalance of attention distribution explains the model's tendency to trust textual information when there is a conflict between textual and visual information.

\subsection{Re-Balancing Contrastive Decoding}

\begin{figure}[!ht]
\includegraphics[width=\textwidth]{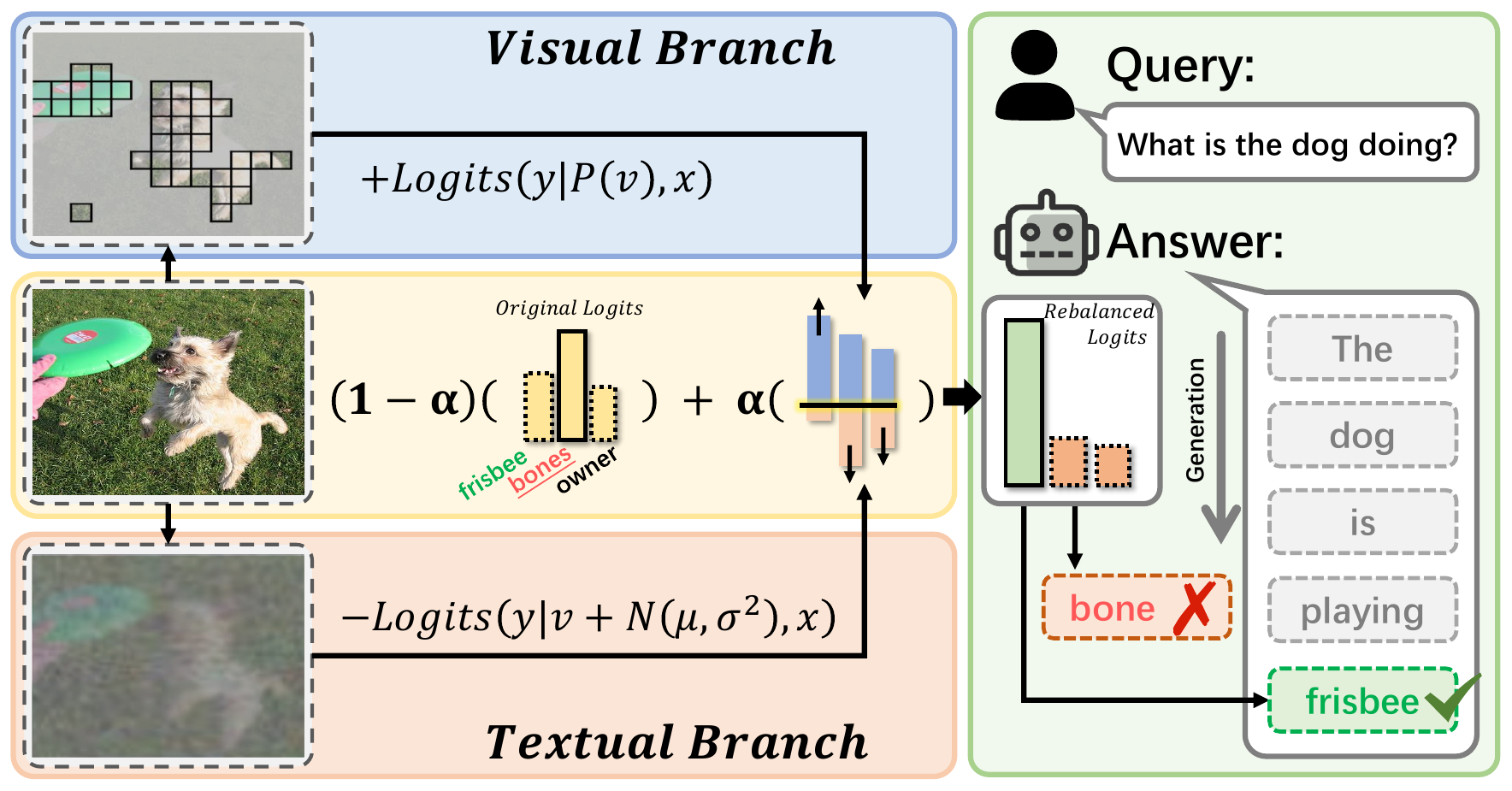}
\caption{Overview of our RBD, which is designed to calibrate the model's preference for textual and visual knowledge in order to mitigate the hallucinations. 
On the left side, logits derived/obtained from \colorbox[RGB]{251,229,214}{textual} and \colorbox[RGB]{222,235,247}{visual} branches are integrated to refine the distribution of \colorbox[RGB]{255,230,153}{original logits} produced by VLM. 
This process amplify the predictions from visual branch while diminishing the untruthful predictions from textual branch, resulting in the final, \colorbox[RGB]{169,209,142}{rebalanced logits} depicted on the right side. 
} \label{figoverall}
\end{figure}

Fig.~\ref{figoverall} presents an overview of our proposed RBD method, which employs two auxiliary branches: a visual and a textual branch. 
By using contrastive decoding, the bias of text knowledge is reduced, and the weight of visual input is increased. 

This approach aims to balance the contribution of textual and visual information, which in turn helps reduce the occurrence of hallucinations.  The balanced model's probabilistic output,\( P_{RBD} \). Formally:   

\begin{equation}
\begin{split}
\label{eq-all}
P_{RBD}(y|v,x) = &
Softmax[
(1- \alpha)logit_{(\theta)}(y|v,x)  \\
&+
\alpha (
logit_{(\theta,v)}(y|v,x) -
logit_{(\theta,t)}(y|v,x)
)]
\end{split}
\end{equation}
\noindent where 
the $logit_{(\theta,v)}$ and $logit_{(\theta,t)}$ are produced by visual and textual branches, respectively. 
The hyperparameter $\boldsymbol{\alpha}$ provides detailed control for adjusting the model's reliance on different sources of information, ensuring a balanced contribution from both modalities. 
The larger the $\boldsymbol{\alpha}$, the more the model is biased towards visual information, which leads to fewer hallucinations.

In Fig.~\ref{figoverall}, the initial model logits, indicated in yellow, demonstrates a predilection for the token ``bones'', despite its absence in the visual input. 
This bias is intensified in the textual branch output, depicted in orange, suggesting that the introduction of noise eliciting the model's inherent biases. Utilizing this phenomenon, we can identify tokens originating from textual biases and subsequently apply penalties to mitigate them. 
And the visual branch, represented in blue, amplifies significant tokens, thus enhancing the visual component's influence.
The effectiveness of using different branches is further explored in the ablation study detailed in Tab.~\ref{tab:chair}.

\subsubsection{Textual Branch} 
By adding noise to the images, we enhance the model's textual preferences, create a negative contrast to identify and mitigate biases. 
Following~\cite{leng2023mitigating}, we add Gaussian noise on the original image, which is considered the most elementary method:  
\begin{equation}
\label{eq-text}
logit_{(\theta,t)} = logit_{\theta}(y|v+\gamma \cdot \mathcal{N} (\mu,\delta ^2 ),x) 
\end{equation}
\noindent Where $\gamma$ donates noise levels on image. To align with the VCD~\cite{leng2023mitigating}, we fix the parameters $\mu$ to 0 and $\delta$ to 1. 

As the noise level rises, the model increasingly depends on the LLM's internal knowledge. Surprisingly, VLMs confidently produce responses even when images are totally noisy or missing, showing over-reliance on textual knowledge. 
This over-reliance often leads to the generation of illusions, demonstrating the model's dependence on textual information.

By intentionally introducing noise, we can stimulate the model's dependence on textual information and observe changes in the logits distribution to identify which tokens are generated based on LLM's internal textual knowledge. Subsequently, we can selectively suppress these tokens to mitigate potential biases introduced by LLMs. 
Further, we try other ways to attenuate images and stimulate the model's dependence on text. The specific effects can be seen in Sec.~\ref{sec-module-wise}.

\subsubsection{Visual Branch} By introducing a token-level attention modulation strategy within Visual Branch, we guide the VLMs to focus on relevant visual tokens. 
\begin{equation}
\label{eq-image}
logit_{(\theta,v)}= logit_{\theta}(y|P(v),x) 
\end{equation}
\noindent Where $P(\cdot)$ refines the attention map by assessing the significance of visual tokens, thereby directing the VLMs towards the more critical visual tokens. Sec.~\ref{sec-module-wise} employs various ablation methods for this purpose.

Firstly, we identify and rank tokens based on their importance~\cite{rao2021dynamicvit,bolya2022token}, focusing on the main subject. This is achieved by computing the cumulative attention score of each visual token in relation to all other tokens. Formally: 
\begin{equation}
\label{eq:importance_score}
s(\text{x}_i) = \frac{1}{N_h} \frac{1}{n} \sum_{h=1}^{N_h} \sum_{j=1}^{n} A^{(h)}(\text{x}_i, \text{x}_j). 
\end{equation} 
\noindent where \( N_h \) represents the number of attention heads, \( n \) is the total number of tokens, and \( A^{(h, l)}(x_i, x_j) \) denotes the attention score from token \( x_i \) to token \( x_j \) at head \( h \). Consequently, token $x_i$ is important if it garners significant attention from the collective tokens across all attention heads.

Secondly, we set our mask according to token significance using the following formula: 

\begin{equation}
\label{eq:masking_operator}
M_A(\text{x}_i) =
    \begin{cases}
    m_i=+1~~&\text{if }s(\text{x}_i) > \theta,\\
    m_i=-1~~&\text{otherwise}.
    \end{cases}
\end{equation}
\noindent where $m_i$ is the corresponding $i$-th column value from matrix $M_A$, which can assume the values $\{-\infty ,-1 ,+1\}$. 
It is note worthy that setting $m_i$ to $-\infty$ indicates discarding the token, thereby optimizing inference speed and conserving computational resources.

Finally, during the model forward inference process, we insert the attention-adjustment matrix mask $M_A$ on all layers. Formally: 
\begin{equation}
\label{eq-beta}
\begin{split}
Softmax(\frac{QK^T}{\sqrt{d_k}} + M_A)&=\frac{ exp(h_i)}{ {\textstyle \sum_{j=1}^{N} exp(h_j)} }=\frac{\beta^{m_i} \cdot exp(k_i)}{ {\textstyle \sum_{j=1}^{N}\beta^{m_i} \cdot exp(k_j)} }  \\
h_i &=  k_i + log(\beta)\cdot m_i  
\end{split}
\end{equation}
\noindent Where $\beta$ represents a scaling factor. the $k_i$ refers to the $i$-th column of query-key attenion score.
Importantly, After passing through the softmax function, this ensures the attention scores are normalized, maintaining an aggregated attention of 1. 
And we add the causal attentional mask $M_C$ to ensure that only previous tokens are concerned when decoding.

This branch heightens the focus on visual tokens within the attention mechanism, directing the model's attention selectively towards or away from specific tokens. Consequently, VLMs can generate responses that better align with the images, reducing potential hallucinations.

\section{Experiments}
\label{sec:exp}

\subsection{Implementation Details}

In this study, we evaluate the effectiveness of our method RBD by using it on three widely-used models: 
LLaVAv1.5~\cite{liu2023improved}, InstructBLIP~\cite{dai2024instructblip} and 
MiniGPT-4~\cite{zhu2023minigpt}. Further experiment details on model architectures, decoding parameters are deferred to Sec.~\ref{sec-hyperp}. 
\subsubsection{Dataset and metric.}

Following prior studies~\cite{li2023evaluating,leng2023mitigating}, we select a random subset of 500 images from the MSCOCO val2014 dataset for evaluating CHAIR and POPE. 
This subset serves as the standardized benchmark for all evaluations. Our assessment employs the following metrics: 
\begin{itemize}[leftmargin=*,nosep]
    \item[$\star$] \textBF{CHAIR}~\cite{rohrbach2018object} (\emph{Caption Hallucination Assessment with Image Relevance}). 

    It employs VLM to generate descriptions and then compares them to the actual objects in the image. The differences are quantified at both the instance (CHAIR$_I$) and sentence (CHAIR$_S$) levels: 
    \begin{gather*}
        \text{CHAIR}_I = \frac{\big|\{\text{hallucinated objects}\}\big|}{\big|\{\text{all mentioned objects}\}\big|}, \\
        \quad \text{CHAIR}_S = \frac{\big|\{\text{captions with hallucinated objects}\}\big|}{\big|\{\text{all captions}\}\big|}.
    \end{gather*}
    
    \item[$\star$] \textBF{POPE}~\cite{li2023evaluating} (\emph{Polling-based Object Probing Evaluation}). 

    It converts hallucination evaluation into a binary classification task by asking whether an object is present in a given image.

    Following POPE~\cite{li2023evaluating}, we report the reslut under the adversarial settting. 
    We assess the performance of VLMs by reporting their accuracy and F1 scores. 
\end{itemize}

\subsubsection{Baselines.} 
\label{sec-basline}

To showcase the inherent capabilities of the model, we utilize greedy decoding as a baseline approach for evaluation. Additionally, we also compare our RBD with other popular methods designed to mitigate hallucinations(Sec.~\ref{sec-3ways2maitigating}), which fall into three categories:

\begin{itemize}[leftmargin=*,nosep]
    \item[$\star$] \emph{\textBF{Original Decoding Strategies}}: 
    Greedy decoding and beam search, which are used to demonstrate the basic performance of the model without additional intervention. 
   
    \item[$\star$] \emph{\textBF{Assistance Based Strategies}}:   
    LURE~\cite{zhou2024analyzing} and Woodpecker~\cite{yin2023woodpecker}, employ supplementary models to alleviate hallucinations or to revise generated descriptions. 
    
    \item[$\star$] \emph{\textBF{Decoding Intervention Strategies}}: 
    Our analysis also extends to recent decoding strategies specifically designed to address hallucinations, such as Contrastive Decoding~\cite{li2023contrastive}, DoLa~\cite{chuang2024dola},  OPERA~\cite{huang2024opera} and VCD~\cite{leng2023mitigating}.  
    
\end{itemize}

\subsection{Comparisons}
\subsubsection{Main Results.}

Our proposed method RBD was benchmarked against previous baseline approaches(Sec.~\ref{sec-basline}) and as indicated by the results in Tab.~\ref{tab:chair}, the VLMs show a marked improvement in mitigating hallucination issues post-rebalancing, particularly with respect to the POPE metric. Moreover, our RBD technique outperformed comparative hallucination mitigation technologies without the need for additional models or tools, highlighting the plug-and-play advantage of our approach. Furthermore, our method consistently surpassed decoding strategies such as CD, DoLa and VCD, further corroborating the robustness and efficacy of our proposed solution in reducing hallucinations.

\begin{table*}[!t]
\caption{\textbf{Comparison of Different Methods Using CHAIR and POPE Metrics.}
The CHAIR metric, where lower scores denote reduced instances of hallucinations. And the POPE metric, where higher scores reflect better performance. 
The highest-performing results are highlighted in \textbf{boldface}, and the second highest are underscored with an \underline{underline} to facilitate a clear comparison. 
}
\label{tab:chair}
\small
\resizebox{\linewidth}{!}{
  \centering
  \begin{tabular}{l|cccc|cccc|cccc}
    \toprule
     \multirow{2}{*}{\textbf{Methods}} 
     & \multicolumn{4}{c|}{\textbf{LLaVAv1.5}} 
     & \multicolumn{4}{c|}{\textbf{InstructBLIP}}  
     & \multicolumn{4}{c}{\textbf{MiniGPT-4}} 
     \\
    & $C_S\downarrow$  &$C_I\downarrow$  &$P_A\uparrow$ & $P_F\uparrow$
    & $C_S\downarrow$  &$C_I\downarrow$  &$P_A\uparrow$ & $P_F\uparrow$
    & $C_S\downarrow$  &$C_I\downarrow$  &$P_A\uparrow$ & $P_F\uparrow$
    \\
    \midrule
Greedy 
&20.8&6.6 &85.9&85.5
&33.1&16.0&79.8&80.3
&28.2&10.7&80.3&80.2\\
Beam Search 
&18.8&6.1 &86.6&85.7
&23.5&8.0 &80.0&80.2
&27.2&10.0&81.5&81.1\\
\midrule
LURE 
&${18.1}$&6.3 &86.8&86.1
&--  &--  &$\underline{80.2}$&$\textbf{81.9}$
&26.1&9.3 &82.2&82.1\\  
Woodpecker 
&$\textbf{17.7}$& 6.4&87.0&86.6
&$\textbf{17.0}$& $\underline{7.2}$&79.0&78.6
&26.0&9.2 &81.5&80.7\\ 
\midrule
OPERA 
&18.3& $\underline{6.1}$&86.8&86.0
&18.3& 7.5&79.6&$\underline{80.8}$
&27.0&10.3&82.7&82.5\\
CD 
&21.6& 6.3&86.5 & 86.4
&24.2& 7.8& -- & -- 
&27.3&10.4& -- & -- \\  
DoLA 
&20.8& 6.5&86.4&86.2
&24.2& 7.8&79.5&79.4
&28.2&10.3&71.8&81.7  \\
VCD 
&20.5& 7.0&87.4&87.0
&21.7& 7.7&79.6&79.5
&27.7&10.8&81.2&81.1\\
\midrule
\rowcolor{blue!15}
RBD w/o textual
&18.8& 6.4&$\underline{87.7}$&$\underline{87.5}$
&18.9 &7.3&79.8 &79.7
&$\underline{25.3}$ & $\underline{8.8}$&$\underline{83.1}$&$\underline{83.0}$ \\
\rowcolor{blue!15}
RBD w/o visual
&20.5& 6.9&87.3 & 87.0
&21.6 &7.6&79.6 & 79.6
&27.5& 10.4&81.2&81.0\\
\rowcolor{blue!15}
RBD 
&$\underline{17.8}$& $\textbf{6.0}$&$\textbf{88.0}$ &$\textbf{87.8}$
&$\underline{17.9}$ &$\textbf{7.2}$&$\textbf{80.4}$ & 80.3
&$\textbf{24.7}$ & $\textbf{8.3}$&$\textbf{84.1}$&$\textbf{84.0}$ \\
\bottomrule
  \end{tabular}
}
\end{table*}

 \subsubsection{Common Benchmarks.}

To further validate whether our RBD impairs the intrinsic capabilities of the original model, we conducted experiments across five widely-used visual question answering benchmarks, including: VQA-v2~\cite{goyal2017making}, GQA~\cite{hudson2019gqa}, VisWiz~\cite{gurari2018vizwiz}, ScienceQA-IMG~\cite{lu2022learn}, TextVQA~\cite{singh2019towards}, and three publicly available general benchmarks, POPE~\cite{li2023evaluating},  MMBench~\cite{liu2023mmbench}, MME~\cite{fu2023mme}. 
The results of these experiments are presented in Tab.~\ref{tab:tongyong}. Our findings indicate that the integration of RBD does not lead to a deterioration in model performance; on the contrary, it even achieves improvements on certain datasets such as POPE and MME. This demonstrates the efficacy of our RBD approach in mitigating VLM hallucinations and enhancing general perceptual abilities.

\begin{table*}[!t]
\caption{\textbf{Comparison among different VLMs on 5 visual question answering benchmarks and 3 common benchmarks.}
Benchmark names are abbreviated due to space limits. The highest-performing results are highlighted in \textBF{boldface}. }
\label{tab:tongyong}
\small
\resizebox{\textwidth}{!}{
  \centering
  \begin{tabular}{lcc|ccccc|ccc}
    \toprule
     \multirow{2}{*}{\textBF{Methods}} & \multirow{2}{*}{\textBF{LLM}} & \multirow{2}{*}{\textBF{Res.}}  & \multicolumn{5}{c|}{\textBF{Visual Question Answering}} & \multicolumn{3}{c}{\textBF{Common }} \\%Benchmark
    &  &  & VQA$^\text{v2}$ & GQA & VisWiz & SQA$^\text{I}$ & VQA$^\text{T}$ & POPE  & MMB & MME  \\
    \midrule
InstructBLIP & Vicuna-13B & 224 & -- & 49.5 & 33.4 & 63.1 & 50.7 & 78.9 & -- & 1212.8 \\

MiniGPT-4 & Vicuna-13B & 224 & 41.0 & 41.0 & 19.6 & 61.0 & 42.5 & 85.3  & -- & 1293.8 \\

LLaVAv1.5 & Vicuna-13B & 336  & \textBF{80.0} & {63.3}& {53.6} & {71.6} & {61.3} & {85.9} & {67.7} & {1531.3} \\    

\rowcolor{blue!15} LLaVAv1.5$_{RBD}$ & Vicuna-13B & 336  & {79.8} & \textBF{63.4} & \textBF{54.0} & \textBF{71.7} & \textBF{61.7} & \textBF{88.3} & \textBF{67.9} & \textBF{1543.3} \\ 
\midrule

\midrule

InstructBLIP & Vicuna-7B & 224 & -- & 49.2 & 34.5 & 60.5 & 50.1 & 79.8  & 36.0 & -- \\

Qwen-VL & Qwen-7B & 448 & \textBF{78.8}& 59.3 & 35.2 & 67.1 & \textBF{63.8} & --  & 38.2 & -- \\
Qwen-VL-Chat & Qwen-7B & 448 & 78.2 & 57.5 & 38.9 & 68.2 & 61.5 & --  & 60.6 & 1487.5 \\
LLaVAv1.5 & Vicuna-7B & 336  & {78.5} & 62.0 & {50.0} & {66.8} & 58.2 & 85.9  & 64.3 & 1510.7 \\

\rowcolor{blue!15}
LLaVAv1.5$_{RBD}$ & Vicuna-7B & 336  & 78.4 & \textBF{62.0} & \textBF{50.8} & \textBF{66.8} & 58.9 & \textBF{88.0}  & \textBF{64.3} & \textBF{1515.8} \\

\bottomrule
  \end{tabular}
  }

\end{table*}

\subsection{Ablation Analysis}

\subsubsection{Decoding Parameters.}

By adjusting one decoding parameters on LLaVAv1.5, while keeping other variables constant, we report the optimal POPE accuracy metric achievable through this parameter adjustment.
Tab.~\ref{tab:ablation_generate} shows that decoding parameters play a critical role in influencing the quality of the generated text. 
Setting the top\_k parameter to 10 make the best accurary of 87.01$\%$. However, to ensure a fair comparison, we employ the simplest greedy search strategy, unless otherwise specified.

\subsubsection{Module-wise Ablation.} 
\label{sec-module-wise}

The effectiveness of several modules was assessed using the POPE metric, as detailed in Tab.~\ref{tab:ablation}. 
For attenuate images and stimulate the model's bias in textual branch, we employed four different approaches: 
(1) without textual branch, 
(1) complete removal of images, 
(2) replacement with image captions, 
(3) introducing noise to images, 
and 
(4) replacement with pure color. 
According to the results presented in Tab.~\ref{tab:ablation_textual}, (3) demonstrates the best performance.

To direct the focus of the visual branch to images, we implemented several strategies: 
(1) Excluding  the visual branch entirely to establish a baseline, 
(2) token pruning based on significance indices, 
(3) token amplification through significance,
(4) enhancing the entire image, 
and 
(5) utilizing referring segmentation to reinforce the emphasis.
Tab.~\ref{tab:ablation_visual}, indicate that strategy (5) yields superior results; however, it necessitates the auxiliary support of other segmentation models. Consequently, we opt for strategy (3), which ranks as the second most effective approach, due to its simplicity.

\begin{table*}[!t]
         \caption{\textbf{Ablation Experiments for LLaVAv1.5-7B Using POPE Metric.} 
         The best performances are highlighted in \textBF{boldface}, and our default settings are marked in gray. }
		\label{tab:ablation}
         \vspace{-0.3cm}

	\begin{center}
                    \subfigure[\textBF{Generation Type}
                    \vspace{-0.3cm}
		              \label{tab:ablation_generate}]{

    						\begin{tabular}{L{2cm}C{1.5cm}}
							\toprule
                                Strategy&\makecell[c]{POPE\\Acc. $\uparrow$}\\
							\midrule
                                \rowcolor{gray!40} Greedy & 85.89\\
                                Beam-2 & 86.51\\
							Beam-5 & 86.67\\
							top\_k &  \textBF{87.01}\\
                                top\_p & 86.52\\
							\bottomrule
						\end{tabular}}
      \hspace{0.01\linewidth}
                    \subfigure[\textBF{Textual Branch}
                        \label{tab:ablation_textual}]{

    						\begin{tabular}{L{2cm}C{1.5cm}}
							\toprule
                                Strategy&\makecell[c]{POPE\\Acc. $\uparrow$}\\
							\midrule
                                w/o & 87.68\\

                                None& 77.61 \\
							Caption-text& 79.29 \\
                                \rowcolor{gray!40} Add noise & \textBF{88.01}   \\
                                Pure color& 78.03\\
							\bottomrule 
						\end{tabular}}
      \hspace{0.01\linewidth}
                  \subfigure[\textBF{Visual Branch}
                        \label{tab:ablation_visual}]{

    						\begin{tabular}{L{2cm}C{1.5cm}}
							\toprule
                                Strategy &\makecell[c]{POPE\\Acc. $\uparrow$}\\
							\midrule
                                w/o & 87.33\\
							Token prune& 87.87 \\
							Amplification& 87.41\\
							\rowcolor{gray!40} Token select & 88.01 \\
                                Referring seg & \textBF{89.12} \\
							\bottomrule
						\end{tabular}}\\
	\end{center}

 \vspace{-0.5cm}
\end{table*}

\subsubsection{Hyper Parameters.} 
\label{sec-hyperp}
\begin{figure}[!ht]
\includegraphics[width=\textwidth]{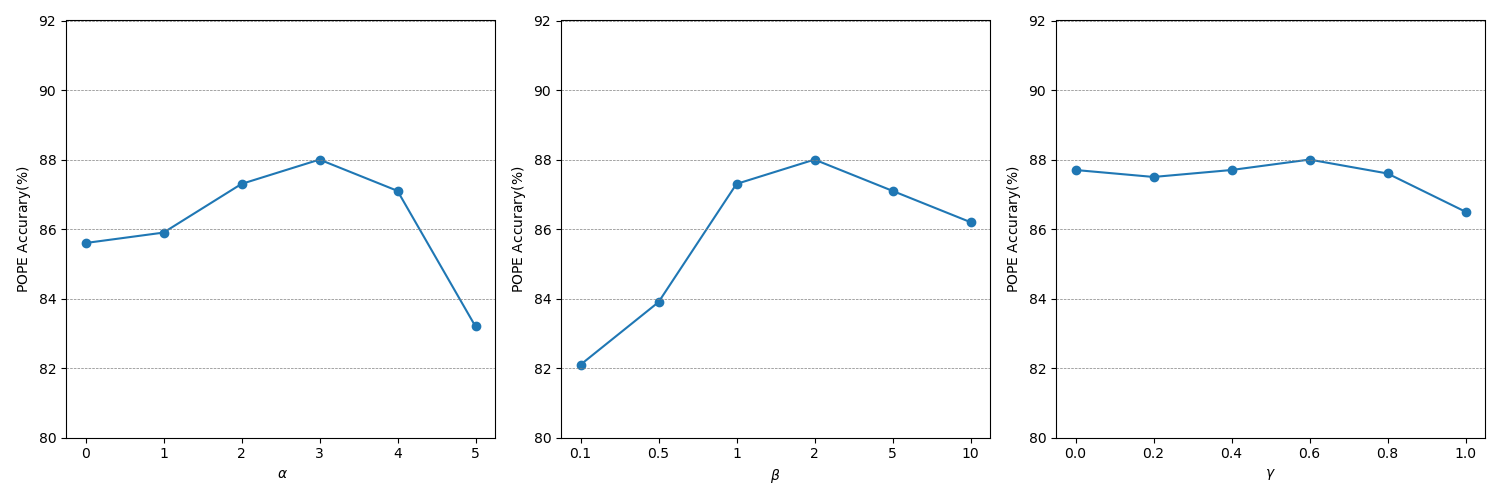}
\caption{
Results when using different hyperparameters on LLaVAv1.5-7B.
Figures show the Accurary metric in POPE. Bigger values indicate fewer hallucinations. 
} 
\label{fig:hyperp}
\end{figure}

Three parameters, $\alpha$ in Eq.~\ref{eq-all}, $\beta$ in Eq.~\ref{eq-beta} and $\gamma$ in Eq.~\ref{eq-text}, may influence the performance of the proposed RBD. Herein, we evaluate the POPE accuracy of LLAVAv1.5-7B. 
Fig.~\ref{fig:hyperp} displayed the POPE accuracy scores for various parameters.
After analysis, the optimal parameter values were determined as $\alpha$ = 0.6, $\beta$ = 2, and $\gamma$ = 0.8, leading to the highest accuracy of 88.0$\%$.

\section{Conclusion}

In this paper, our RBD method addresses the issue of Multimodal Knowledge Conflicting Hallucinations in VLMs. 
By incorporating auxiliary branches, RBD rebalances the weight between textual and visual information during inference, enhancing the VLMs' fidelity to visual content without the necessity for extensive model restructuring or additional computational resources. 
Experimental results demonstrate a marked reduction in hallucinations and improved accuracy, suggesting that RBD paves the way for more sophisticated multimodal integration. 
Future research could explore the interplay between attention mechanisms and modality alignment to uncover a more efficacious method.

\subsubsection{Acknowledgments}
This work is supported by the National Natural Science Foundation of China (Grant No. U21B2004) and the Zhejiang Provincial Key RD Program of China (Grant No. 2021C01119).

\bibliographystyle{splncs04}
\bibliography{Main}

\end{document}